# A deep-learning view of chemical space designed to facilitate drug discovery

Paul Maragakis,[1,*,†] Hunter Nisonoff,[1,*] Brian Cole,[1] and David E. Shaw[1,2,†]

[1] D. E. Shaw Research, New York, NY 10036, USA.

[2] Department of Biochemistry and Molecular Biophysics, Columbia University,

New York, NY 10032, USA.

[*] These authors contributed equally to the manuscript.

† To whom correspondence should be addressed.

Paul Maragakis

E-mail: Paul.Maragakis@DEShawResearch.com

Phone: (212) 478-0414

Fax: (212) 845-1414

David E. Shaw

E-mail: David.Shaw@DEShawResearch.com

Phone: (212) 478-0260

Fax: (212) 845-1286

## Abstract

Drug discovery projects entail cycles of design, synthesis, and testing that yield a series of chemically related small molecules whose properties, such as binding affinity to a given target protein, are progressively tailored to a particular drug discovery goal. The use of deep learning technologies could augment the typical practice of using human intuition in the design cycle, and thereby expedite drug discovery projects. Here we present DESMILES, a deep neural network model that advances the state of the art in machine learning approaches to molecular design. We applied DESMILES to a previously published benchmark that assesses the ability of a method to modify input molecules to inhibit the dopamine receptor D2, and DESMILES yielded a 77% lower failure rate compared to state-of-the-art models. To explain the ability of DESMILES to hone molecular properties, we visualize a layer of the DESMILES network, and further demonstrate this ability by using DESMILES to tailor the same molecules used in the D2 benchmark test to dock more potently against seven different receptors.

**Introduction**

Small-molecule drug discovery is typically an iterative process in which existing knowledge is used to design new molecules for synthesis and testing, resulting in a series of molecules with incrementally improved properties. The process of drug discovery is expensive and time-consuming, and its duration is mostly determined by the number of cycles of design, synthesis, and testing needed to improve the properties of successive generations of small molecules to match the various criteria needed in a drug discovery project. Deciding how to modify small molecules in each iteration typically relies on human intuition, and interest is growing in developing machine learning models that can learn from existing data to propose molecules with modifications that are more likely to improve properties of interest. Such properties might include the strength of interaction with a drug discovery target or anti-target, or any of a number of properties (e.g., hydrophobicity or stability) related to small-molecule administration, metabolism, and pharmacokinetics. If machine learning models that propose improvements to small molecules can help scientists make better decisions during the design stage, they could potentially make drug discovery programs considerably more efficient and likely to succeed.[1–3]

There exist a number of different representations of small molecules, and each representation has certain advantages and disadvantages for the purposes of molecular design.[1] The most common choice of representation is a so-called "molecular graph," which establishes the covalent connectivity of the atoms and any specified stereochemistry (or lack thereof in the case of racemic mixtures). All graphs can be represented explicitly by sets of vertices and edges; this particular representation, however, has historically been hard to manipulate effectively using deep learning methods (although progress in deep learning of generic graphs is currently underway[4–7]). A simplified representation specifically for small-molecule graphs is SMILES, a compact text-based representation in which chemical species are mapped to single ASCII strings of 20–90 characters;[8] this representation has been used successfully in deep learning models that generate molecules.[1,3,9–13] A SMILES string is advantageous because it encodes an exact structural representation of a molecule, but molecules with only small chemical differences can have very different SMILES strings, making them poor representations of chemical similarity and thus difficult to use to tailor molecular properties.

Another representation, a so-called "extended connectivity fingerprint,"[14] encodes the local chemical environment around a molecule's heavy atoms. Chemically similar molecules generally have similar fingerprints, and this advantage over the SMILES representation has led to fingerprints being commonly used as the molecular representation in computer codes that evaluate chemical properties, or that search for similar molecules in a database.[15] As opposed to SMILES, however, a fingerprint is a structurally lossy encoding of a molecule: A given fingerprint does not always correspond one-to-one with a single molecule, and furthermore the task of generating molecular structures from a fingerprint has historically been a significant problem, making it infeasible to take advantage of the ability to tailor the chemical properties of fingerprints in order to develop improved molecules.

We reasoned that a deep learning model could be developed to translate fingerprints to SMILES, and that additional fine-tuning of such a model could make it effective for proposing modified molecules that are both structurally legible and have improved properties of interest. (We note that a recent preprint[16] demonstrated the feasibility of producing compatible SMILES strings from a given fingerprint, but modifying the model to propose molecules with improvements to particular properties was outside the scope of that work.)

Here we present DESMILES, a deep learning model that advances the state of the art in machine learning approaches to molecular design. DESMILES operates by generating a series of small molecules that are chemically related to a given small-molecule input; in its most basic form, DESMILES takes a molecular fingerprint as input and translates the fingerprint to a sequence of SMILES strings in order of an estimated probability of matching the fingerprint, starting with the most likely one. In its typical use, the DESMILES network is modified with transfer learning (learning of downstream tasks with additional training on small, specialized datasets), and then proposes a sequence of molecules that are chemically similar to the input but likely improved for specified properties of interest. (Inspired by the recent successes of models developed for analogous problems in natural language processing (NLP),[17–23] we drew on state-of-the-art ideas in NLP when designing the neural network architecture.)

To teach DESMILES to produce molecules that are useful in drug discovery, we trained it on synthetically feasible molecules (drawn from the eMolecules, molPort, and SureChEMBL chemistry libraries) that were developed prior to 2017 and that had already passed a number of computational filters commonly used in drug discovery, such as removing molecules that are likely to be promiscuous, unstable, or reactive. To demonstrate that DESMILES can generate synthesizable and nontrivially novel molecules, we show that it recovered from their fingerprints over 94% of all molecules that passed the computational filters and were patented in 2017 or later (i.e., at a later date than molecules used in the training set).

We demonstrate that DESMILES succeeds on a number of downstream drug discovery tasks. Specifically, we applied DESMILES to a previously published benchmark[24] that assesses the ability of a method to modify input molecules to inhibit the dopamine receptor D2, and DESMILES yielded a 77% lower failure rate compared to the state of the art,[25] and yielded a 4% lower failure rate compared to the state of the art for improving so-called "molecular beauty." Furthermore, we demonstrate that DESMILES can successfully modify an input molecule to dock more potently against any given receptor, by further training the model using results from docking a moderately sized library of commercial molecules to the receptor. To help explain why DESMILES is successful in downstream discovery tasks, we analyze an internal layer of the trained model and find that it inherits a desirable organization of chemical space from the fingerprints: Visually the layer is reminiscent of an energy landscape,[26] with nearby basins corresponding to chemically similar molecules and vice versa. Transfer learning gradually deforms this "chemical landscape," biasing the basins to molecules better suited for the downstream drug discovery task of interest.

## Results

DESMILES is a recurrent neural network architecture that takes an input fingerprint and produces a stream of output molecules ranked by an estimated probability that their fingerprint matches the input fingerprint (Methods). We designed DESMILES by performing a hyperparameter scan over various architectural elements and selecting the model that maximized the number of molecules in our validation set that were correctly found from their fingerprints. The validation set consisted of molecules from the SureChEMBL library that were published more recently than those in the training set, and we evaluated the final DESMILES model on a test set consisting of SureChEMBL molecules published in April 2018 (Methods). We found that our best single model recovered 94.1% of the validation molecules and 92.8% of the test molecules. When we trained the model with the optimal architecture on the combined training and validation sets, it recovered 94.84% of the test molecules.

Our so-called "ablation" study (Extended Data Table 1) suggests that regularization and the global optimization algorithm, A*, are key ingredients in our optimal architecture. Using the optimal architecture, we found that the top ranked molecule generated by A* was likely to exactly match the input fingerprint, and subsequent molecules generated by A* had fingerprints similar to the input fingerprint (Figure 1a). A composite model constructed from twenty slightly different architectures recovered 97.96% of the validation set and 97.41% of the test set; adding the model trained on the combined training and validation sets, the composite model recovered 98% of the test set. For the 2% of molecules that were not recovered, the correct canonical SMILES had a very low estimated probability for all the individual models. (Extended Data Table 2 shows the importance of the value of the estimated probability in determining whether or not a molecule was recovered.)

To gain insight into how DESMILES organizes chemical space, we visualized the internal chemical landscape derived from an internal layer of the neural network (Figures 1b–d). The surface of the top A* solution is reminiscent of an (inverted) energy landscape[26] in which neighboring basins contain chemically similar molecules. The surfaces of the second, third, and subsequent high-ranking solutions chosen by A* display patterns similar to those of the top solution. A molecule that is highly ranked in one region of this chemical landscape will decay to lower ranks as the distance from that region increases. Distances in the chemical landscape correlate with distances in fingerprint space, as expected (Extended Data Figure 1). Formal tools for exploring energy landscapes have been used successfully in protein folding and in the study of atomic clusters,[26] and related sampling methods can be used to explore models that generate molecules.[27] We found, however, that direct transfer learning by applying additional training on specialized datasets was a simple and effective way to leverage the organization of chemical space learned by the pre-trained DESMILES neural network, and so we took that approach here.

As initial examples of this transfer learning approach, we used DESMILES to propose inhibitors of the dopamine receptor DRD2, and to propose molecules with a high quantitative estimate of drug-likeness (QED),[28] based on existing data.[24] For each of these tasks, we trained DESMILES on a dataset consisting of matched molecular pairs (pairs of similar molecules A and B, where A is the input molecule and B is the output molecule with an improved property), and evaluated the performance of the model on two benchmark metrics in comparison to five existing deep learning models and a conventional matched-pair analysis algorithm.[24] When each model was permitted to generate twenty prospective molecules (as specified in the benchmark), the failure rate for obtaining a successful inhibitor of DRD2 was 4.4 times lower with DESMILES (3.2%) than with the previous state of the art,[25] the hierarchical graph-to-graph translation method (14.4%).[24] Our model remained better, even with only three prospective molecules, than the previous state of the art with twenty prospective molecules (Figure 2a). Using a composite

DESMILES model and allowing more than twenty prospective molecules enabled us to achieve a failure rate as low as 0.2% (Extended Data Figure 2). On the QED benchmark metric, the failure rate for producing at least one molecule (among twenty prospective molecules) with a better QED score than that of the input molecule was 4% smaller with DESMILES (22.1%) than with the hierarchical graph-to-graph translation (23.1%).[25] (Additional details are provided in Extended Data Table 3. Results from an additional benchmark test, LogP, are given in Extended Data Table 4.) The strong performance of DESMILES on these benchmark tests suggests that the model successfully captures fundamental aspects of chemical space.

To understand why the model performed well, we examined how the pre-trained neural network parameters gradually transformed during transfer learning for the DRD2 benchmark test. We found that the internal chemical landscape deformed as the parameters changed, such that groups of similar molecules migrated together as more potent molecules were pushed up to the surface, until the majority of the local landscape was occupied by molecules predicted to be potent inhibitors of DRD2 (Figure 2b). Similar transformations occurred throughout the chemical landscape at slightly different rates, and six so-called "epochs" (passes of the DRD2 training set through the neural network) were required for optimization (Extended Data Figure 3). The limited nature of the additional training (in terms of the number of examples and epochs, and the similarity of the matched pairs) allowed the network to keep chemically similar molecules close together while also favoring better-scoring molecules representing incremental modifications.

We reasoned that these reorganizations of groups of neighbors in the DESMILES chemical landscape could also apply in transfer learning of other diverse small molecule properties. A frequently encountered task in drug discovery is to use a 3D model of a drug target to discover new small molecules that bind to that target, and this can be viewed as a docking problem. We evaluated the ability of DESMILES to improve the docking scores of random scaffolds from weak binders against seven targets from the enhanced directory of useful decoys (DUD-E)[29] (genes: ADA, ALDR, CAH2, COMT, CP2C9, TRY1, TRYB1). For each receptor, we constructed a training set by identifying matched pairs containing a potent binder and a chemically similar but weak binder from the original DESMILES training set (Methods). We found that the seven resulting DESMILES models each proposed neighboring molecules with significantly improved docking scores for their corresponding receptor (Figure 3a). The modifications DESMILES made to the input molecule were diverse and specific to both the receptor and the input molecule (Figure 3b and Extended Data Figure 4) and reminiscent of the way in which medicinal chemists transfer structural elements between scaffolds.

**Conclusions**

Here we have presented DESMILES, a deep learning model that has learned chemistry from real molecules relevant for drug discovery, that exploits the observation that similar molecules have similar properties, and that can learn to transfer useful chemical patterns to new applications concerning small molecules, as we have shown. Our example applications suggest that DESMILES can be used as an idea generator to support medicinal chemists. We observed improved performance when the training set was expanded to reflect additions to the patent literature, suggesting that DESMILES can be kept up to date. Additional modifications—such as restricting the chemical landscape to molecules accessible from available starting materials for synthesis,[30] or training on libraries specific to a class of drug targets—could further increase the value of DESMILES in certain drug-discovery contexts. Although the main examples we discussed in this paper involved making improvements to existing molecules, DESMILES could also be used early in the drug-discovery process: By expanding on the transfer learning strategies described here, DESMILES could potentially be combined with emerging molecular screening technologies[31,32] and long-timescale molecular dynamics simulations[33] to streamline the process of identifying new scaffolds for drug candidates. Researchers in other industries concerned with small molecules may also be able to adapt DESMILES for a variety of applications.

## Methods

*Data and preprocessing*

To develop DESMILES, we used three collections of drug-like molecules: 1) the MolPort database[34] of over seven million purchasable compounds (in January 2017); 2) the eMolecules database[35] of over six million purchasable compounds (in October 2015); and 3) the SureChEMBL database[36] of molecules extracted from the chemical literature that contained over 16 million molecules (in March 2018). All molecules were extracted in the simplified molecular-input line-entry system (SMILES) format[8,37] and converted to a canonical SMILES representation using RDKit.[38] In order to retain only drug-like molecules, we applied a number of filters to the raw data. First, we removed all molecules that contained more than 70 heavy atoms. Next, we filtered molecules that contained an atom that was not in the set {H, C, N, O, F, P, S, Cl, Br, I} and applied several in-house filters that removed compounds predicted to have undesirable properties, such as reactivity and mutagenicity. Finally, we used MolVS[39] to remove salts, by selecting the largest fragment, and to neutralize the molecules.

*Tokenization*

We tokenized the SMILES with SentencePiece[40] using the byte-pair-encoding (BPE) algorithm,[17] keeping 4 tokens (<start>, <forward>, <reversed>, and <end>) for the start, left-to-right direction, right-to-left direction, and end of a SMILES string. The total number of tokens is a hyperparameter; after measuring the recovery of molecules from the validation set using 2,000, 4,000, and 8,000 tokens, we decided to use 8,000 tokens. With this choice 99.96% of the training molecules could be represented by strings of 27 or fewer tokens (the 5th, 50th, and 95th percentiles had 6, 9, and 14 tokens respectively). We removed all SMILES strings that consisted of more than 27 tokens.

*Fingerprints*

DESMILES takes as inputs concatenated ECFP4 and ECFP6 fingerprints.[14] The ECFP4 and ECFP6 fingerprints are computed using RDKit and include chirality bits. Both fingerprints use 2,048 bits, such that the total input to the model is 4,096 bits. The percentage of training molecules with 4,096 bits exactly matching another training molecule was 0.12%. We "recovered" a molecule when the fingerprint of the generated molecule exactly matched the input fingerprint. We found that in 1.04% of the test molecules, the first generated molecule that recovered the fingerprint had a different small molecule graph; typically DESMILES also generated the exact input molecule further down in the stream of generated molecules, except for 0.06% of molecules (six out of 10,000), which it did not generate within our typical fixed compute time.

***Training, validation, and test set construction***

In order to be useful as a tool in drug discovery, an algorithm must be able to generalize beyond the areas of chemical space that it has seen. We hypothesized that a random split of the data would be insufficient to measure the ability of DESMILES to generalize to new areas of chemical space, since many molecules in the test set would be close neighbors of molecules in the training set, and we thus constructed a validation set using a temporal split. We constructed the training set from MolPort,[34] eMolecules,[35] and SureChEMBL[36] molecules published before 2017. The validation set consists of a random subset of 10,000 molecules from SureChEMBL published between January 1, 2017 and January 16, 2018. The test set consists of a random subset of 10,000 molecules published by SureChEMBL in April of 2018. During the initial development of the DESMILES architecture, we used a random 80% subset of the training set and tested the performance of our best model using the remaining 20% of the training set (finding it recovered 99.72% of these molecules). Finally, we trained a model with the optimal architecture on the full training set.

*Algorithm details*

*Architecture*

The main inspiration for the DESMILES architecture comes from advances in "sequence-to-sequence" models, such as those found in neural machine translation,[41] and corresponding adaptations of these approaches to "vector-to-sequence" problems, such as those found in image captioning.[42] The DESMILES neural network is trained to maximize the likelihood, $p(Y|f)$, of producing a target sequence of tokens, $Y = (y_1, y_2,...)$, that correspond to the SMILES string of the molecule from which the fingerprint, f, was generated. Common neural machine translation architectures consist of "encoder" and "decoder" networks. The encoder network is tasked with learning a semantic representation of the source language sentence. When the encoder is a recurrent neural network (RNN),[43] the semantic representation is typically the final hidden state of the encoder RNN. This final encoder hidden state is then used to initialize the decoder hidden state. The decoder is then tasked to generate the target sequence from the information provided by the final encoder hidden state. Similarly, for image captioning, an encoder network (typically a convolutional neural network) learns a fixed-length representation of the image, which is then fed into a decoder RNN to generate a caption.

Here, rather than encoding a sequence of tokens or an image, DESMILES encodes the molecular fingerprint. The encoding of the fingerprint is then passed to a decoder RNN, which learns to generate the corresponding SMILES string. A schematic of the architecture is provided in Extended Data Figure 6. Since the fingerprint can be represented as a fixed-length vector, we used a simple two-layer feed-forward neural network with tanh activation functions for the encoder. The size of the output of the final layer was restricted to be the same size as that of the decoder hidden state. Each layer of the encoder consists of batch normalization, an affine transformation, and a hyperbolic-tangent activation function. During training, dropout was applied to the second layer after the batch normalization with probability 0.1.

The decoder consists of weight-dropped long short-term memory networks (LSTMs),[18,44] which apply a DropConnect[45] mask to the hidden-to-hidden recurrent weights during training as a recurrent regularization technique. We adjusted an implementation of the average-stochastic gradient descent weight-dropped (AWD) LSTM.[46] The hidden and cell states of the weight-dropped LSTM of the first layer are initialized to the output of the encoder network. When predicting the next token in the target sequence during training, all previous tokens of the target sequence are passed through an embedding layer before passing as input to the decoder LSTM (by the so-called "teacher forcing" method). In addition to DropConnect, we borrowed a number of regularization techniques for LSTMs[18] that were found to provide substantial improvements in training language models. These techniques, which are outlined below, are (1) embedding dropout, (2) variational dropout, (3) weight tying, and (4) activation regularization and temporal activation regularization.

During training, embedding dropout[47] was used, wherein dropout is applied simultaneously to all the elements of the token's vector embedding. The remaining embeddings are scaled by $(1-p_e)^{-1}$, where $p_e$ is the probability of the embedding dropout. In addition, variational dropout,[47] in which the same dropout mask is applied at every time step, was used in multiple places during training. First, with probability $p_i$, an individual element of the token embedding was selected for variational dropout before being passed as input to the LSTM. Variational dropout was then applied to the outputs of every layer of the LSTM, with the exception of the final layer, with probability $p_h$. Finally, variational dropout was applied to the outputs of the final layer of the LSTM with probability $p_e$. The input size of the LSTM of the first layer and the output size of the LSTM of the final layer are constrained to match the size of the embedding dimensionality. The outputs of the final layer of the LSTM are fed into a "softmax" layer consisting of a linear transformation followed by a softmax activation function. We used weight tying, in which the weights of the embedding and softmax layers are shared.[48,49]

We considered three hyperparameters governing the architecture of DESMILES: the output size of the embedding matrix for the tokens, the size of the hidden state of the LSTM, and the number of LSTM layers. The sizes of the hidden layers of the fingerprint encoder were fixed at the size of the hidden state of the LSTM. The hyperparameters were chosen with a grid search by evaluating the performance on the validation set. This led to the decision to use embedded tokens of dimension 400, an LSTM hidden state of dimension 2,000, and five stacked LSTM layers.

*Loss function and regularization*

The loss function consists of the sum of the negative log-likelihood of the correct token at each step. In addition, activation regularization (AR) and temporal activation regularization (TAR)[18] are applied to the outputs of the final layer of the LSTM. AR and TAR are applied with their own scaling coefficients to weight their contribution to the total loss. Finally, weight decay is applied, in which the weights of the network are multiplied by a factor less than one after each update. We used a scaling coefficient of two for AR, a scaling coefficient of one for TAR, and a multiplicative factor of $10^{-7}$ for weight decay.

*Training*

We trained the model with the Adam optimizer[50] using the "1cycle" learning rate and momentum schedule.[23] Gradient norms were clipped at 0.3. We considered three hyperparameters concerning that training schedule: the number of epochs, the maximum learning rate, and the so-called "dividing factor." The initial learning rate is calculated by dividing the maximum learning rate by the dividing factor. The final learning rate is calculated by dividing the initial learning rate by the square of the dividing factor. During the first 49% of the training schedule, the learning rate increases linearly from the initial learning rate to the maximum learning rate while the momentum decreases linearly from 0.8 to 0.6. Over the next 49% of the training schedule, the learning rate decreases linearly from the maximum learning rate to the initial learning rate, while the momentum increases linearly from 0.6 to 0.8. In the last 1% of training, the momentum is fixed at 0.8, while the learning rate decreases linearly from the initial learning rate to the final learning rate. To train DESMILES, we used a dividing factor of 10. We chose the maximum learning rate and number of epochs using a grid search, with performance evaluated on a validation set. The optimal hyperparameters were 128 epochs and a maximum learning rate of $10^{-3}$. This corresponds to an initial learning rate of $10^{-4}$ and a final learning rate of $10^{-6}$.

During training, we chose batch sizes to maximize the number of samples that could fit on a single GPU (either a GeForce GTX 1080 Ti or GeForce GTX 1080). This number was typically between 150 and 300. During training, we used teacher forcing.[51] To encourage DESMILES to learn to translate the fingerprint into both the forward (left-to-right) and reversed (right-to-left) SMILES string, we randomly reversed half of all SMILES strings in each batch. The second token (following the start token) is a special token that specifies whether or not the SMILES string is reversed.

*Decoding with A\*, a branch-and-bound algorithm*

In our model, the probability of a sequence of tokens $y_1,\ldots,y_T$ given a fingerprint, f, is given by $p(y_1,\ldots,y_T|f) = \prod_{t=1}^{T} p(y_t|f, y_1,\ldots,y_{t-1})$. A common technique used in neural machine translation to search for the most probable target sentence is the beam search algorithm,[52] which is a greedy algorithm. Here, we used a branch-and-bound algorithm to provably enumerate a gap-free list of SMILES strings in order of probability. A combinatorial search tree is constructed as follows: (1) The root node consists of the START token; (2) the children of any node represent all possible tokens at the next position of the byte-pair encoded SMILES string; and (3) a leaf node is any node with an END token. The algorithm enumerates in order of probability the paths in the tree that start with the root node and end in a leaf node. The particular branch-and-bound algorithm we used was the A* algorithm.[53] In order to enumerate SMILES strings in order of probability, we searched for the SMILES string that minimizes the negative log-probability specified by DESMILES. In brief, the algorithm recursively selects the highest-probability partial path from a priority queue, then calculates the probability for all the children of the selected partial path. For any partial path in the tree, x, we must specify a bound, f(x), on the cost function that we wish to minimize. In the A* algorithm, we write f(x) = g(x) + h(x), where x is a node in the search tree, g(x) is the cost up to the node x, and h(x) is an admissible heuristic that bounds the remaining cost to any leaf node under x. We compute g(x) as the sum of the negative log of the conditional probabilities emitted by DESMILES for the tokens specified by the path from the root node to x. Here, h(x) is a lower bound on the cost of the remaining path to any leaf node under x. Since we can bound any negative log conditional probability by 0, we use h(x) = 0 for all x, which is equivalent to Dijkstra's shortest path algorithm for a tree with weighted edges. Although in theory the network could learn arbitrary length sequences, in this study we restricted ourselves to sequences of at most 27 tokens. This covered 99.96% of the post-processed SMILES strings from our databases. To enforce this restriction, any node at the end of a path of 27 tokens is considered a leaf node and no further children are expanded.

To guarantee an answer within a fixed amount of computational effort, we used a modified form of the A* algorithm, which, after branching 5,000 times, returns a leaf node from the priority queue until at most 10,000 leaf nodes are reached. Since the algorithm for generating leaf nodes is deterministic, we returned only leaf nodes that correspond to valid molecules that were not previously enumerated during the search. We used RDKit[38] to check the validity of SMILES strings. If the SMILES string is reversed, it is flipped to its forward representation before processing with RDKit.

### *Recovery of molecules that match a fingerprint*

For each leaf node enumerated, the molecular fingerprint from the SMILES string is computed using RDKit[38] and checked against the input. The search ends after a SMILES string is found that corresponds to a molecule with a matched fingerprint or after 10,000 leaf nodes are enumerated. Our trained DESMILES model finished the search in fewer than 5,000 branches for 93.02% of the 10,000 validation molecules; the median was 16 branches and the 75th percentile was 32 branches.

*Transfer learning for property optimization*

Here the goal was to use DESMILES to take as input the fingerprint of an existing molecule and generate new molecules that (1) were reasonably similar to the starting molecule as specified by a similarity threshold using the Tanimoto similarity[54] to the ECFP4 fingerprint of the molecules and (2) had an improved property (e.g., better predicted biological activity against the dopamine type 2 receptor, DRD2). This was accomplished by assembling a new specialized training set consisting of pairs of molecules, such that the second molecule of each pair satisfied conditions (1) and (2). We then fine-tuned the pre-trained DESMILES network by additional training on this set, effectively asking the network to write the SMILES of the second molecule in the pair, given the fingerprint of the first.

*Fine-tuning DESMILES for improved biological activity against DRD2, drug likeness, and logP*

We used the training, validation, and test sets provided in the paper by Jin et al.,[24] and used the "1cycle" learning rate and momentum schedule described above. We tested the performance on the validation set of 50 random combinations of three hyperparameters: We picked between five and twelve epochs (inclusive), a maximum learning rate from the set {0.002, 0.001, 0.0005}, and a dividing factor between five and ten (inclusive). We performed a grid search around the most promising combination of hyperparameters and selected the best-performing parameters for evaluation of the test set.

We first found the best-performing hyperparameters for testing biological activity against DRD2. For this test, we found that the optimal hyperparameters were six epochs, a maximum learning rate of 0.001, and a dividing factor of seven. We next selected hyperparameters for testing the quantitative estimate of drug-likeness (QED). For this test, we found that the optimal hyperparameters were nine epochs, a maximum learning rate of 0.0005, and a dividing factor of eight. Finally, we selected hyperparameters for testing the penalized octanol-water partition coefficient (logP). For the LogP test with a similarity constraint of 0.6, the optimal hyperparameters were 10 epochs, a maximum learning rate of 0.0005, and a dividing factor of 10. For the LogP test with a similarity constraint of 0.4, the optimal hyperparameters were five epochs, a maximum learning rate of 0.0005, and a dividing factor of six. For each molecule in the test set, DESMILES generated 20 molecules using the A* algorithm described above. Details of how performance is measured can be found in the paper by Jin et al.[24]

To understand the gradual transformation of chemical space during fine-tuning in the DRD2 example (Figure 2b and Extended Data Figure 3), we used the DRD2 optimal hyperparameters, and additionally, during training we froze the layer that maps the fingerprint to a continuous vector (everything to the left of the orange circle in Figure 1b) to anchor the input molecules in the portion of embedded space that we visualized, so that we could see the gradual transformation of the output molecules starting from the same input molecules.

*Fine-tuning DESMILES for improved docking scores*

We fine-tuned DESMILES to generate a set of molecules with improved docking scores for each of seven different receptors: ADA, ALDR, CAH2, COMT, CP2C9, TRY1, and TRYB1. We performed all docking using an in-house method. We generated training pairs from the original DESMILES training set, excluding all molecules in the DUD-E set for each receptor, as follows. 1) We randomly selected 50,000 molecules and docked them. 2) We took the top 5% of molecules and, for each of them, found up to 50 molecules with an ECFP4 similarity of at least 0.4. 3) We docked those additional molecules and kept only those with a docking score weaker than their parent molecule by a determined threshold. A threshold of 30 units was used for CP2C9, TRY1 and TRYB1. A threshold of 60 units was used for ADA, ALDR, CAH2, and COMT in order to keep the number of training pairs comparable between the seven receptors. For the transfer learning, we used the hyperparameters chosen for the DRD2 experiment as above. We evaluated the fine-tuned DESMILES using the molecules from the DRD2 validation set: For each molecule, DESMILES generated 20 candidate molecules. All generated molecules with an ECFP4 similarity of at least 0.4 to the starting molecule were docked against the corresponding receptor.

## Acknowledgements

The authors thank Michael Eastwood for helpful discussions and Jessica McGillen and Berkman Frank for editorial assistance.

# Figures

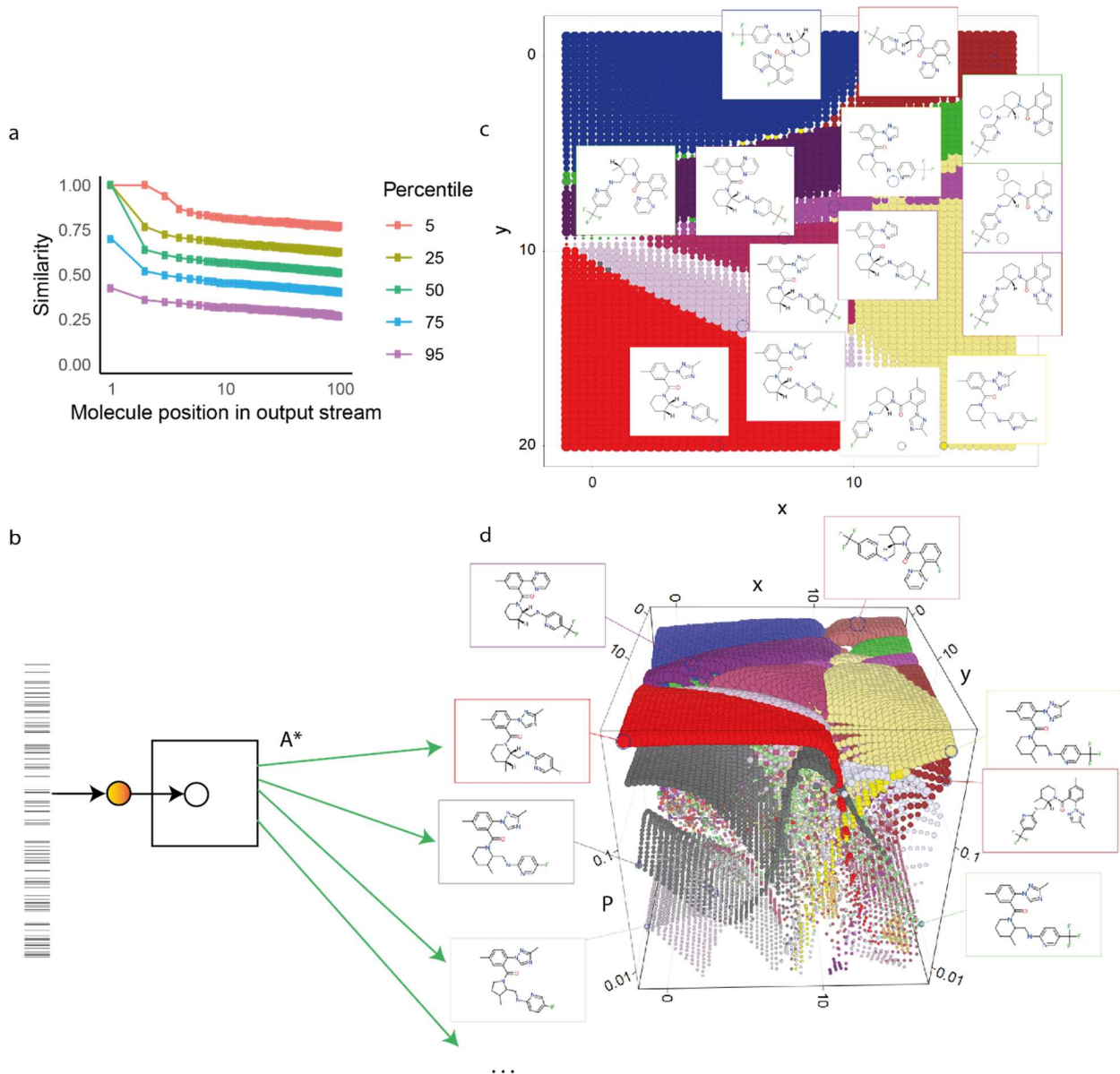

**Figure 1. Generating molecules with DESMILES. a)** Similarity of the molecules generated by DESMILES to the input fingerprints for the validation set. DESMILES outputs an ordered stream of molecules for each input fingerprint, and the x-axis denotes the position in the stream

(1 being the first solution found by the A* algorithm, 2 the second, etc.). The colored curves show, as a function of position in the output stream, the percentile of validation molecules for which the Tanimoto similarity to the output molecule (measured by Tanimoto similarity of their ECFP4 fingerprints) is at least the value on the y-axis. **b)** Schematic of the conversion of fingerprints to a stream of molecules by DESMILES. The input fingerprint is converted to a 2,000-dimensional embedding (orange circle), then passed through another layer to the internal layers of the neural network (square; see Methods for details), where it initializes the hidden and so-called "cell" states (empty circle). The A* solutions form a ranked stream of generated molecules (indicated by green arrows). **c)** Probability surface of the top A* solution as a function of coordinates in a cut of the embedding space that spans three neighboring molecules in the validation set (the x- and y-axes define the plane of those three molecules). Connected regions with circles of the same color correspond to the same molecular structure, with a few representative structures shown. The size of each circle is proportional to the probability DESMILES estimates for the top A* solution at that point in chemical space. **d)** The DESMILES chemical landscape. 3D scatterplot of the probability surfaces of the top five A* solutions (the x- and y-axes are the same as in panel **c**, and the z-axis shows the probability, P, on a logarithmic scale). The probability of the top A* solutions decreases as they approach a crossing point where the original molecule switches to a neighboring molecule.

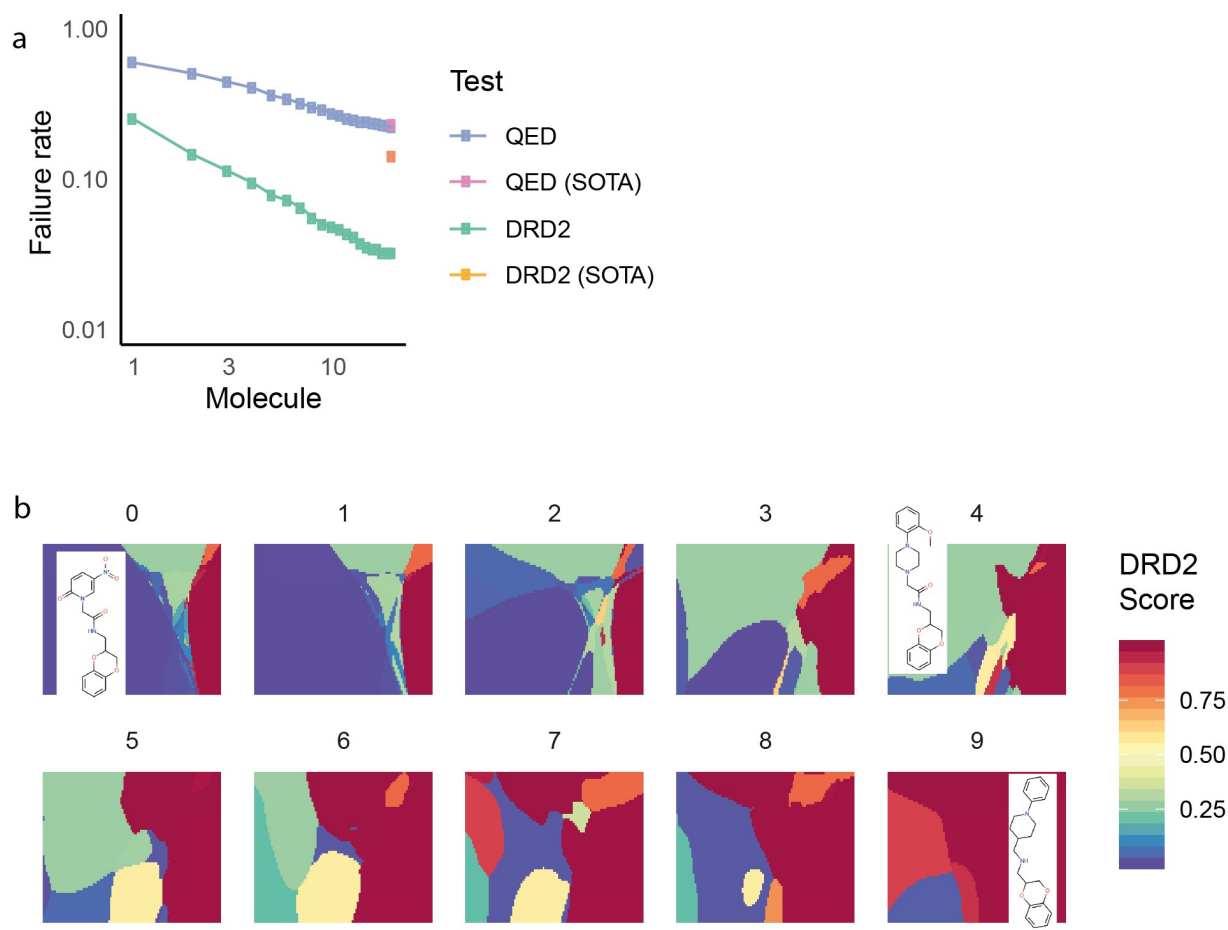

**Figure 2. Transfer learning applied to benchmarks in lead optimization. a)** The failure rate for DESMILES to generate a better neighbor in terms of drug likeness (QED) or predicted inhibition of the dopamine receptor D2 (DRD2), as a function of the number of molecules generated from each test molecule (on a log-log scale). The yellow and pink data points show the previous state of the art (SOTA) for the DRD2 and QED benchmarks, respectively. **b)** The gradual transformation of the DESMILES chemical landscape during the DRD2 transfer learning process. The chemical landscape near a molecule from the test set (highlighted in panel 0) and two molecules from the training set (highlighted in panels 4 and 9) is shown. The color code shows the likelihood that the first molecule produced (i.e., the top A* solution) from a coordinate of the embedding layer will be an inhibitor of DRD2. Each panel covers the same part of a 2D

plane in the embedding layer. Panel 0 is from the pre-trained model before transfer learning. Each successive panel is a snapshot taken during transfer learning, over the course of which the network sees 4,000 examples of pairs of molecules with one member of the pair having improved DRD2 activity. The ten panels cover one so-called "epoch" of training. In panels 2 through 5, a small modification of a non-potent test molecule migrates toward the region previously occupied by that test molecule; in panels 6 through 9, the modified molecule is then pushed away by a more potent molecule that includes an additional modification in the linker.

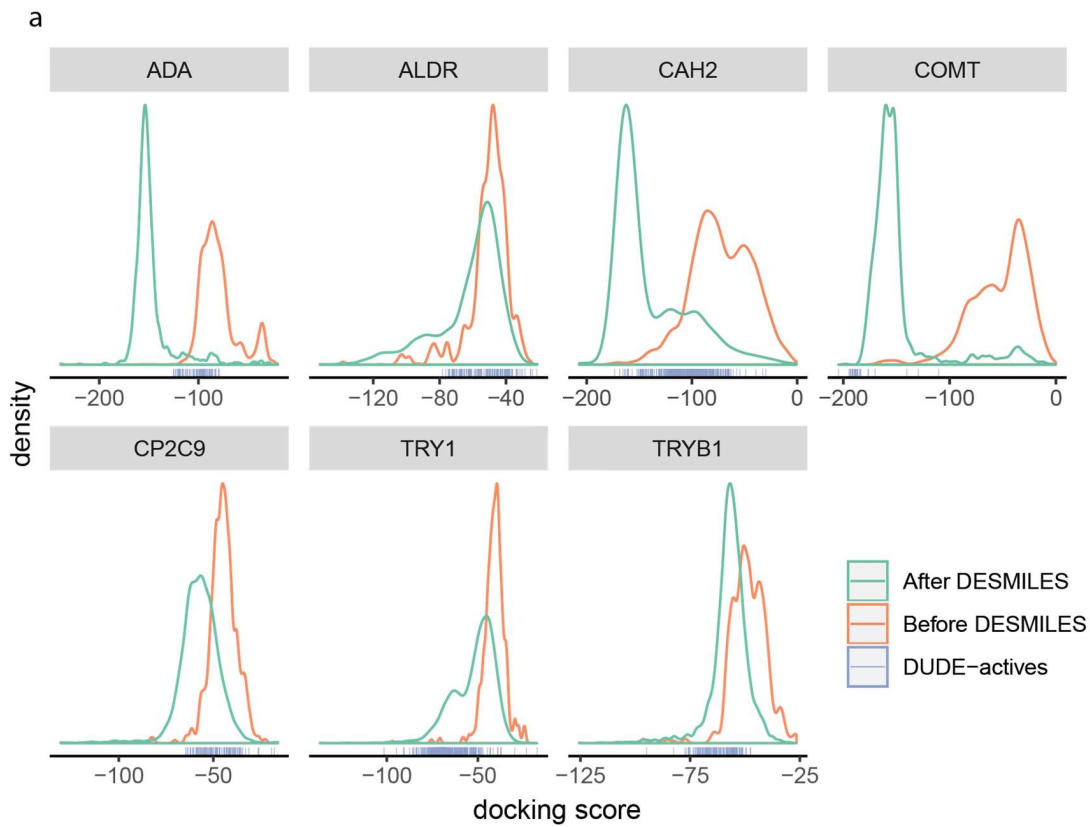

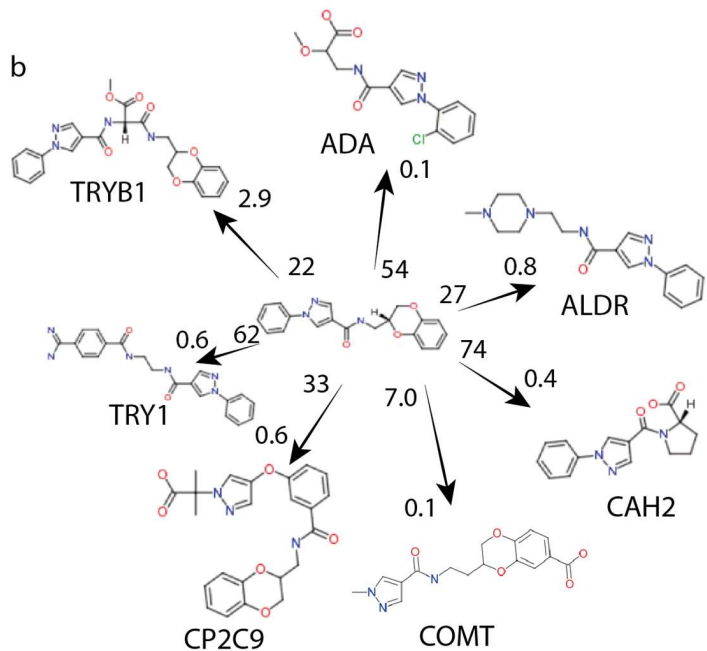

**Figure 3. Transfer learning of docking. a)** Distributions of docking scores before and after

applying a DESMILES model that was trained on each of seven receptors from the DUD-E data set;[29] the "rug" plot under each pair of distributions shows the docking scores for the set of DUD-E actives. More negative docking scores indicate stronger predicted binding. The number of training examples of potent molecules per target varied from 210 molecules for ALDR to 4,697 for ADA, with a corresponding variation in the shift of the distributions of docking scores. The number of potent molecules per target was: ADA, 210; ALDR, 4,697; CAH2, 4,423; COMT, 3,898; CP2C9, 2,179; TRY1, 905; TRYB1, 1,503. **b)** A molecule from the test set and seven neighbors generated by DESMILES, each of which docks well against one of seven different receptors. The arrows show the percentile ranking of the starting molecule (at arrow bases) and the produced molecule (at arrowheads) compared to the ranking of 50,000 random molecules.

# A deep-learning view of chemical space designed to facilitate drug discovery: Extended Data

## Extended Data Tables

| Model variant | Number of molecules recovered from validation set |
|---|---|
| DESMILES (With A*) | 9,360 |
| **Data augmentation** | |
| No reversed SMILES | 9,247 |
| Only reversed SMILES | 9,100 |
| **Regularization techniques** | |
| No AR/TAR | 9,339 |
| No DropConnect | 9,345 |
| No weight tying | 9,189 |
| No weight decay | 8,757 |
| No DropConnect, no dropout | 9,229 |
| No DropConnect, no dropout, no weight tying | 9,083 |
| **Inference techniques** | |
| Beam search with beam width 1 | 6,088 |
| Beam search with beam width 10 | 8,230 |
| Beam search with beam width 100 | 8,965 |
| Random sampling with 100 tries | 8,395 |
| Random sampling with 1,000 tries | 8,914 |

**Extended Data Table 1.** Results from our ablation study, in which we tested the performance of variants of the DESMILES model (rows) on our validation data set. Except for the changes listed in the table, we used the hyperparameters of the optimal model architecture (see Extended

Data Table 5) and optimal training schedule (Methods). The full model selected by the A* algorithm (top row) significantly outperformed all model variants employing other inference techniques. Particularly important factors influencing model performance were weight decay, weight tying, and using both forward and reversed SMILES. The validation data set contained a total of 10,000 molecules.

| Variable | Scaled variable importance |
|---|---|
| Probability | 148.5 |
| FSP3 | 16.7 |
| MW | 14.1 |
| XLogP | 13.8 |
| AromaticRingCount | 13.6 |
| BondCount | 13.4 |

**Extended Data Table 2.** Importance of variables in a random forest classifier[1] that predicted whether or not molecules from the validation data set were recovered by the optimal DESMILES model. The variables we considered for the classification were 28 common molecular properties, calculated using Vortex,[2] as well as the probability of the SMILES string given its fingerprint ("probability"). Shown are this probability and the five most important molecular properties for the classification. When the probability was less than $5 \times 10^{-13}$, the correct molecule was never recovered, and when the probability was larger than $5 \times 10^{-5}$, the correct molecule was always recovered. The random forest classifier was implemented using the randomForest package in R.[3]

|              | QED     |           | DRD2    |           |
|-------------:|--------:|----------:|--------:|----------:|
|              | Success | Diversity | Success | Diversity |
| MMPA         | 32.9%   | 0.236     | 46.4%   | 0.275     |
| JT-VAE[4]    | 8.8%    | -         | 3.4%    | -         |
| GCPN         | 9.4%    | 0.216     | 4.4%    | 0.152     |
| VSeq2Seq     | 58.5%   | 0.331     | 75.9%   | 0.176     |
| VJTNN+GAN    | 60.6%   | 0.376     | 78.4%   | 0.162     |
| AtomG2G      | 73.6%   | 0.421     | 75.8%   | 0.128     |
| HierG2G      | 76.9%   | **0.477** | 85.9%   | 0.192     |
| DESMILES     | **77.8%** | 0.412   | **96.8%** | **0.328** |

**Extended Data Table 3.** Comparison of methods (rows) using the QED (left panel) and DRD2 (right panel) benchmark tests. The values in the first 5 rows are from Table 2 in the paper by Jin et al.[5] The rows AtomG2G and HierG2G are from Table 1 in the paper by Jin et al.[6] Here, "success" is the success rate of the learned task (either to improve drug likeness, QED, or to identify strong binders to the dopamine receptor DRD2), and "diversity" is the average pairwise Tanimoto distance among the molecules generated from the benchmark data set. The best results are denoted in bold.

|          | Similarity 0.6 | | Similarity 0.4 | |
|----------|---------------|---------|---------------|---------|
|          | Improvement   | Diversity | Improvement | Diversity |
| MMPA     | 1.65 ± 1.44   | 0.329   | 3.29 ± 1.12   | 0.496   |
| JT-VAE   | 0.28 ± 0.79   | -       | 1.03 ± 1.39   | -       |
| GCPN     | 0.79 ± 0.63   | -       | 2.49 ± 1.30   | -       |
| VSeq2Seq | 2.33 ± 1.17   | 0.331   | 3.37 ± 1.75   | 0.471   |
| VJTNN    | 2.33 ± 1.24   | 0.333   | 3.55 ± 1.67   | 0.480   |
| AtomG2G  | 2.41 ± 1.19   | 0.379   | **3.98** ± 1.54 | 0.563 |
| HierG2G  | **2.49 ± 1.09** | **0.381** | **3.98** ± 1.46 | **0.564** |
| DESMILES | 2.43 ± 1.11   | 0.299   | 3.04 ± 1.36   | 0.408   |

**Extended Data Table 4.** Comparison of methods (rows) using the LogP benchmark test. The values in the first 5 rows are from Table 2 in the paper by Jin et al.[5] The rows AtomG2G and HierG2G are from Table 1 in the paper by Jin et al.[6] Here, "improvement" is the improvement in the penalized logP score of molecules under a similarity constraint of 0.6 (left panel) or 0.4 (right panel), and "diversity" is the average pairwise Tanimoto distance among the molecules generated from the benchmark data set at the given level of similarity. The best results are denoted in bold.

|  | **Size** (m rows and/or n columns in vector or matrix) | **Probability** |
|---|---|---|
| **Encoder** | | |
| Batch normalization | m = 4086 | |
| Linear layer with tanh activation | m = 4086, n = 2000 | |
| Batch normalization | m = 2000 | |
| Dropout | | 0.1 |
| Linear layer with tanh activation | m = 4086, n = 2000 | |
| | | |
| **Decoder** | | |
| Embedding layer with embedding dropout ($p_e$) | m = 8000, n = 400 | 0.014 |
| Variational dropout ($p_i$) | | 0.175 |
| Weight dropped RNN layer | m = 400, n = 2000 | 0.14 |
| Variational dropout ($p_h$) | | 0.105 |
| Weight dropped RNN layer | m = 2000, n = 2000 | 0.14 |
| Variational dropout ($p_h$) | | 0.105 |
| Weight dropped RNN layer | m = 2000, n = 2000 | 0.14 |
| Variational dropout $p_h$) | | 0.105 |
| Weight dropped RNN layer | m = 2000, n = 2000 | 0.14 |
| Variational dropout ($p_h$) | | 0.105 |
| Weight dropped RNN layer | m = 2000, n = 400 | 0.14 |
| Variational dropout ($p_o$) | | 0.07 |
| Linear layer (weights tied to embedding layer) | m = 400, n = 8000 | |

**Extended Data Table 5.** Hyperparameters (left column) describing the optimal DESMILES architecture. Shown are the size of the vector or matrix (middle column) and/or probability (right column) associated with each hyperparameter. There were 134,515,392 trainable parameters in total.

# Extended Data Figures

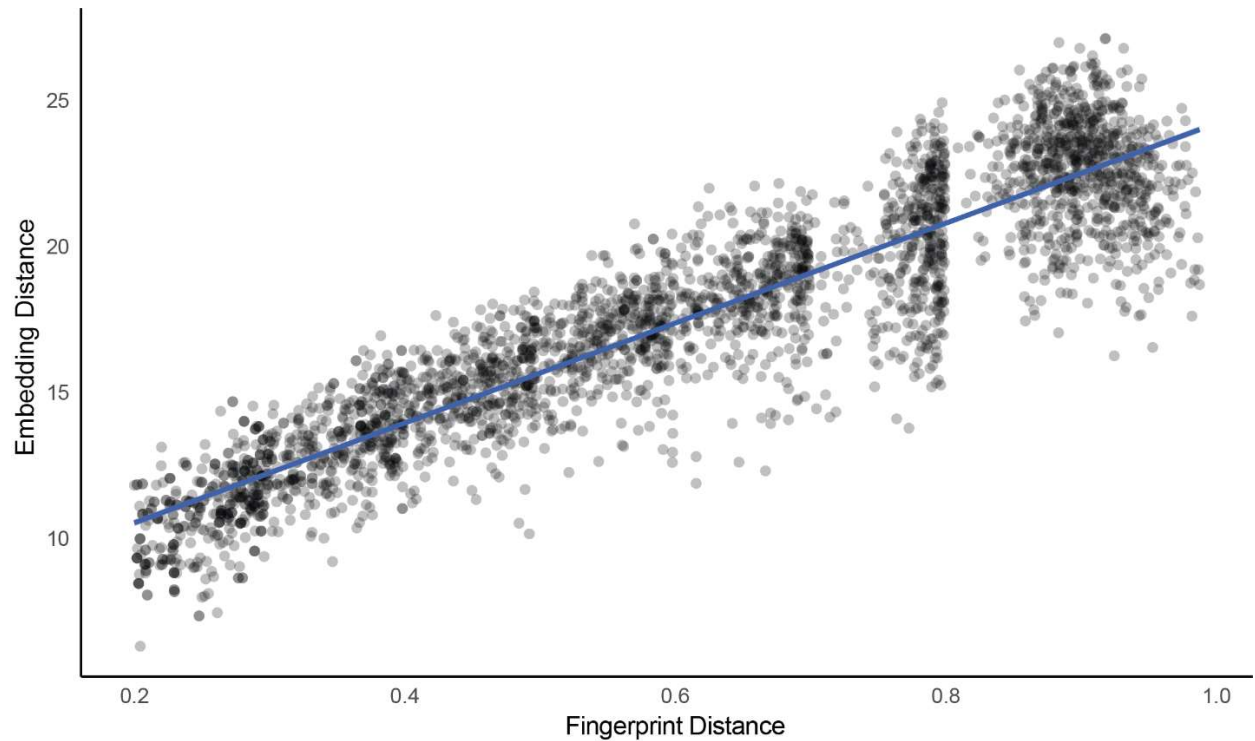

**Extended Data Figure 1.** The original fingerprint distance (x-axis) correlates strongly with the embedded fingerprint distance (y-axis). To examine the relationship between the Jaccard distances of the fingerprints and the Euclidean distances in the embedding space, we binned all pairs of the 10,000 validation fingerprints by their Jaccard distance in bins of width 0.1. We then sampled 400 pairs from each bin and computed their distances in the embedding space. This gave a Pearson correlation coefficient of 0.92.

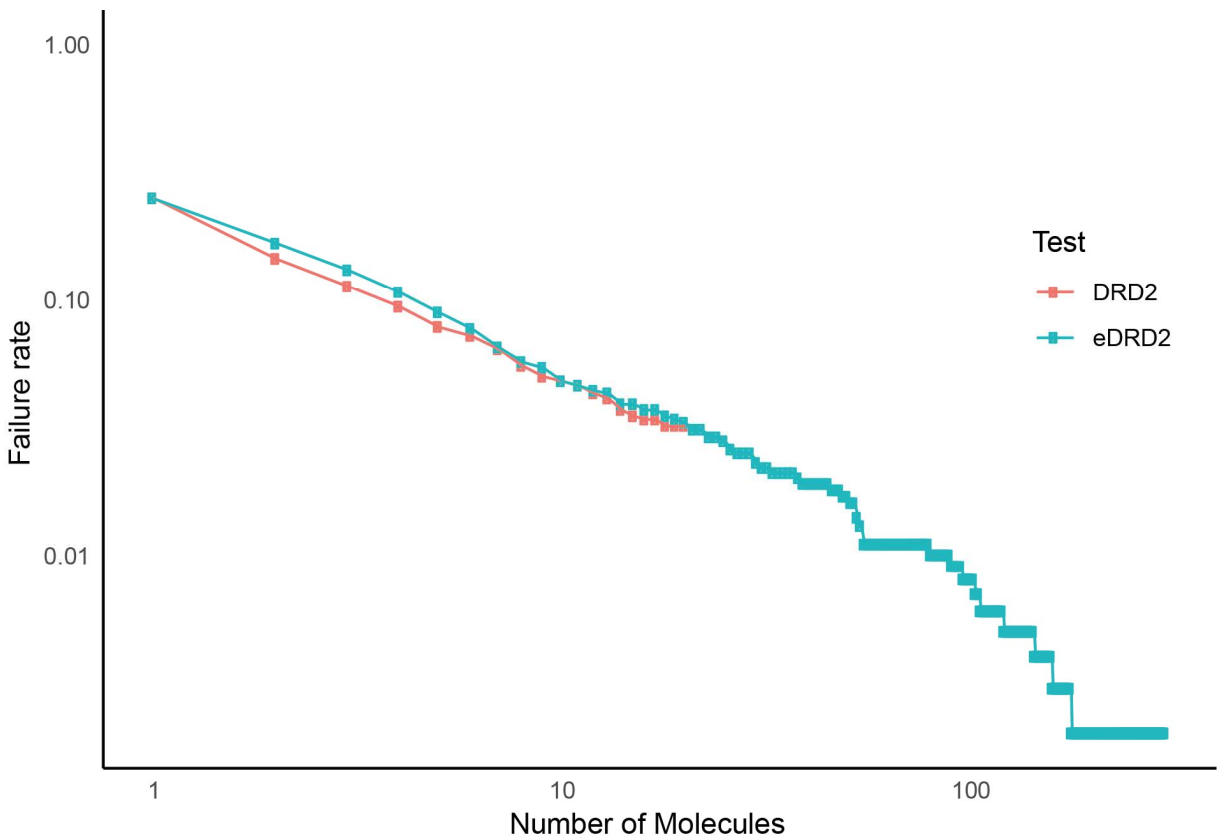

**Extended Data Figure 2.** Failure rate (y-axis) of DESMILES to take an input test molecule and generate an output stream consisting of a given number of neighboring molecules (x-axis) that contained a predicted inhibitor of the dopamine receptor DRD2. We show results for both a single DESMILES model (orange) and an ensemble of 20 DESMILES models (blue). For the latter, the molecule at each position in the overall output stream was sampled by selecting the most probable molecule at that position in the constituent models' streams.

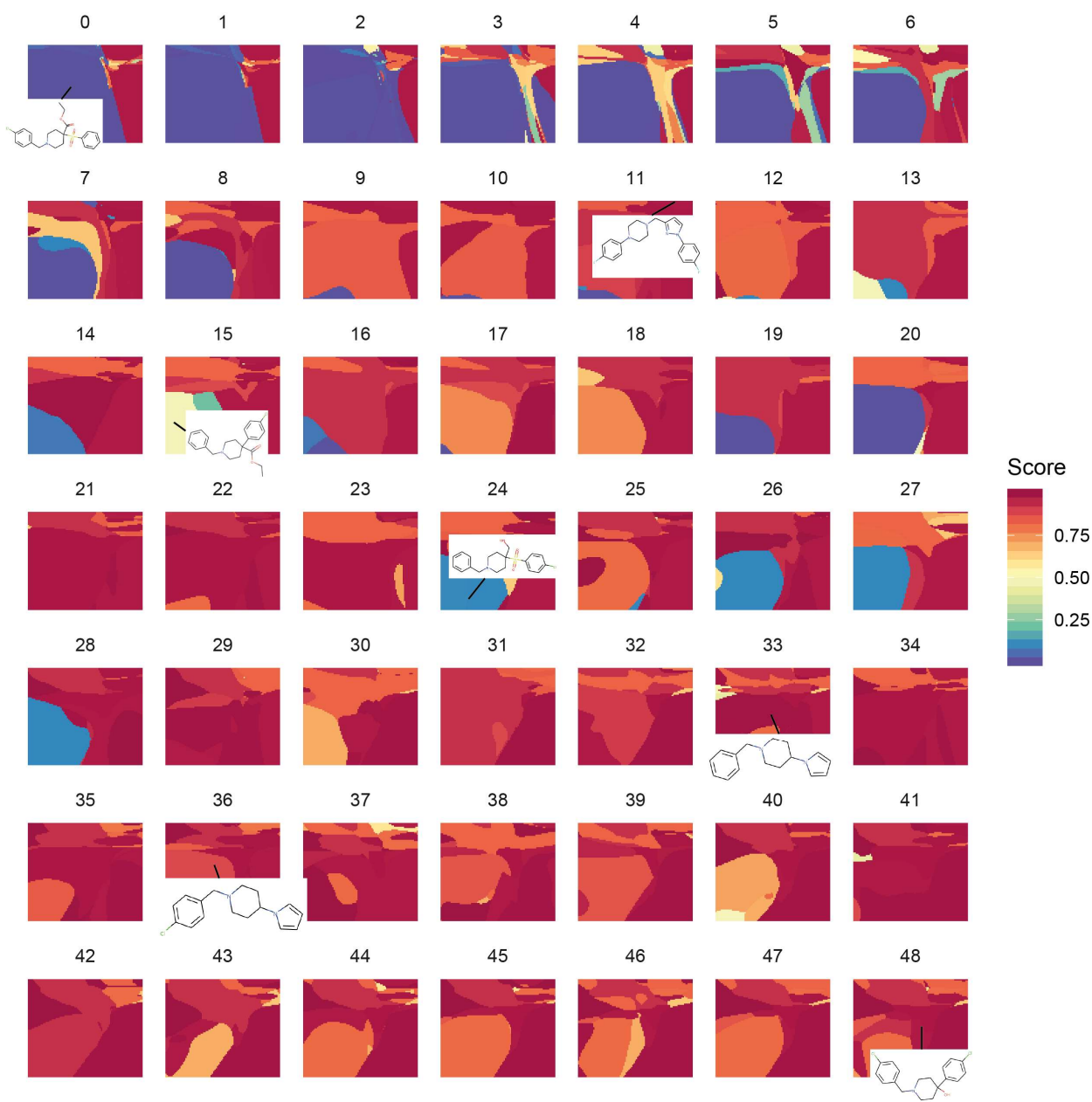

**Extended Data Figure 3.** Gradual transformation, during transfer learning of the DRD2 inhibitor function, of the chemical landscape of the top A* solutions near a molecule of the test set (highlighted in panel 0). Each panel covers the same part of a 2D plane in the embedding layer that contains the test set molecule and two newly generated molecules shown in panels 33

and 48. Panel 0 uses the pre-trained model before transfer learning. The panels interpolate through six epochs of transfer learning. Panels are colored according to the likelihood that the first molecule produced from a coordinate of the embedding layer will be an inhibitor of DRD2. In the center panel (panel 24), the learning rate of the one-cycle learning is maximal and introduces some new molecules to the chemical landscape; these molecules become further optimized during the final half of the one-cycle learning.

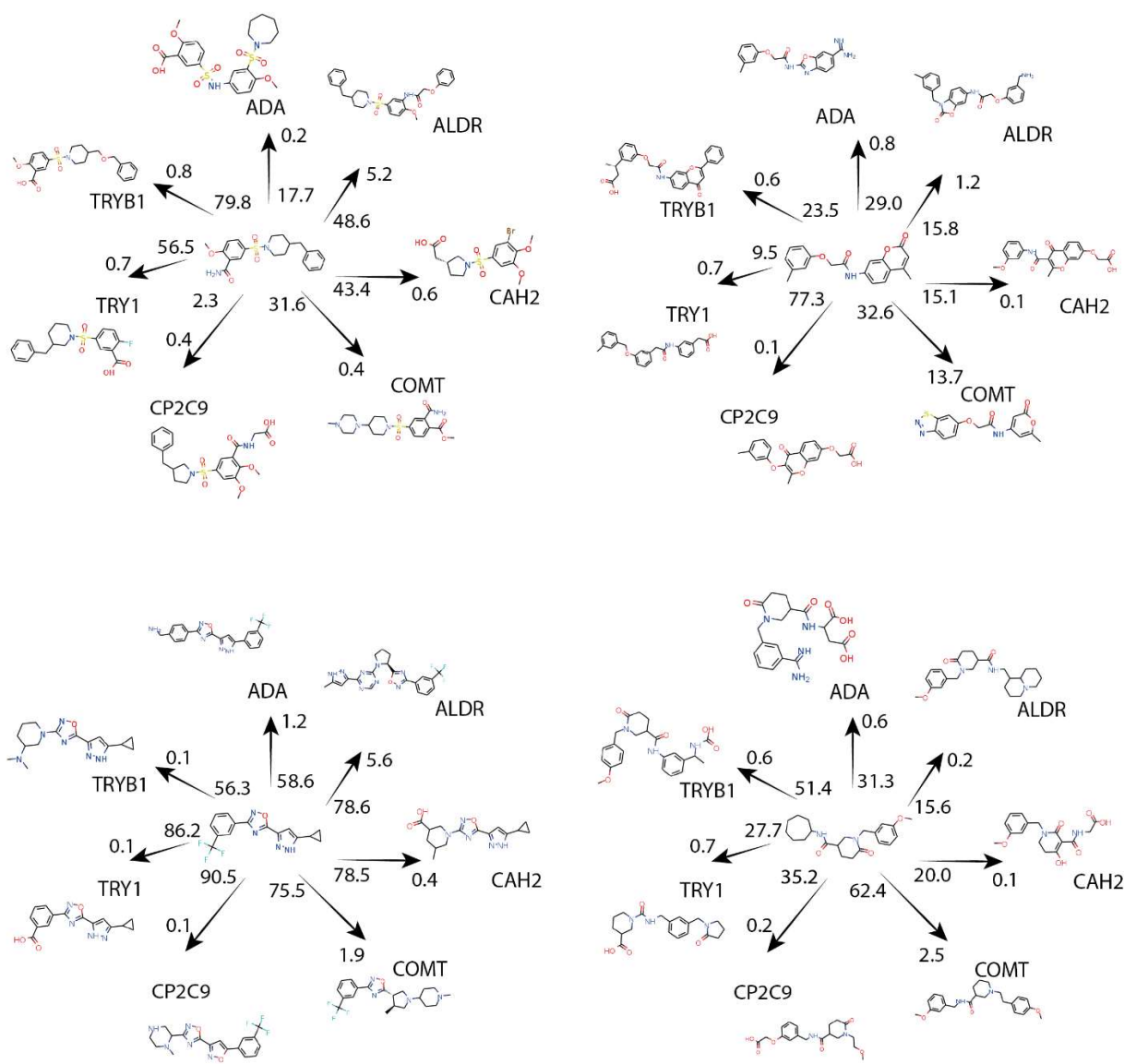

**Extended Data Figure 4.** Examples of diverse modifications to four different molecules that allow them to dock potently against each of seven different receptors. As in Figure 3b, the arrows show the percentile ranking of the starting molecule (at arrow bases) and the produced molecule (at arrowheads) compared to the ranking of 50,000 random molecules.

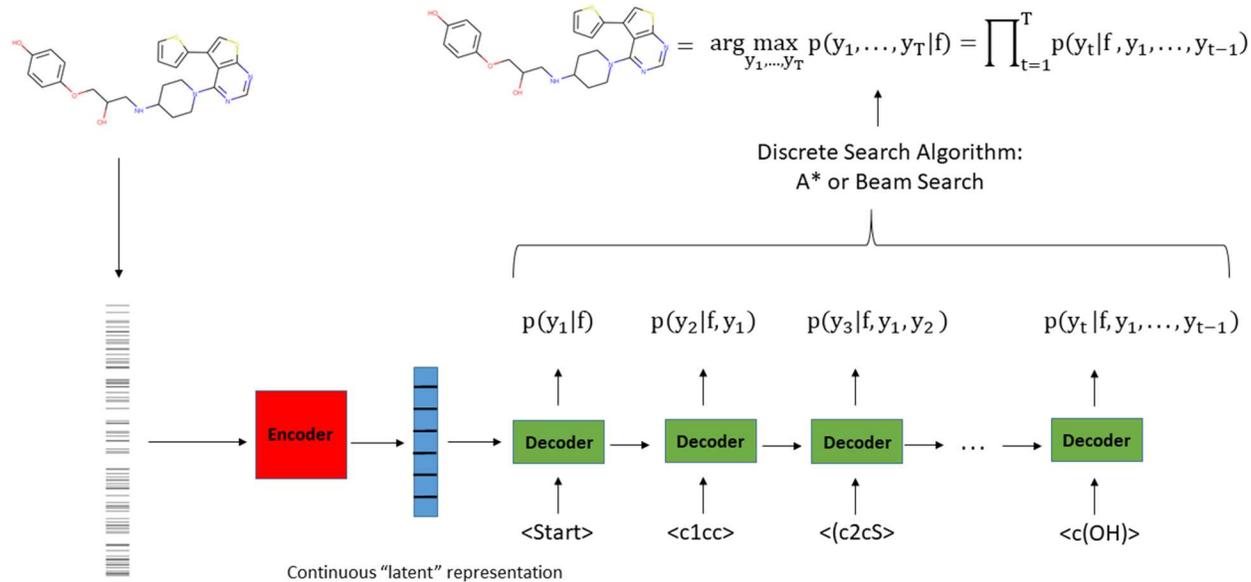

**Extended Data Figure 5.** Schematic of the DESMILES architecture. An input molecule is converted to its extended connectivity fingerprint (top left), which is then passed through an encoder that converts it to a continuous latent representation. The recursive application of the decoder (green) generates probabilities for writing additional "chunks" of the output SMILES. The sequence of output chunks is controlled by a search algorithm that attempts to optimize the estimated probability of the resulting output SMILES.

## Extended Data References